%% file: root.tex
\documentclass[letterpaper, 10 pt, conference]{ieeeconf}  %

\IEEEoverridecommandlockouts                              %

\title{\LARGE \bf
Seeing Farther and Smarter: Value-Guided Multi-Path Reflection \\for VLM Policy Optimization
}

\author{Yanting Yang$^{1}$ \quad Shenyuan Gao$^{2}$ \quad Qingwen Bu$^{3}$ \quad Li Chen$^{3\dagger}$ \quad Dimitris N.Metaxas$^{1}$
\thanks{$^{1}$Rutgers University; $^{2}$The Hong Kong University of Science and Technology; $^{3}$The University of Hong Kong; $^{\dagger}$Corresponding author.}}%
\usepackage{algorithm}
\usepackage{algorithmic}
\usepackage{bbm}
\usepackage{etoolbox}
\usepackage{multicol}
\usepackage{cite}
\usepackage[bookmarks=true]{hyperref}
\usepackage{graphics} %
\usepackage{times} %
\usepackage{amsmath} %
\usepackage{amssymb}  %
\usepackage{svg}
\usepackage{mathrsfs}
\usepackage{amsfonts}
\usepackage{wrapfig}
\usepackage{kantlipsum}
\usepackage{subcaption}
\usepackage{marvosym}
\usepackage{graphicx}
\usepackage{float}
\usepackage{ctable}
\usepackage{cuted}
\usepackage{colortbl}
\usepackage{multirow}
\usepackage{wrapfig}
\usepackage{xcolor}   %
\usepackage{pifont}    %
\usepackage{bm}
\usepackage{caption}
\usepackage{setspace}
\let\labelindent\relax
\usepackage{enumitem}
\usepackage{threeparttable}
\usepackage{varwidth}

\begin{document}

\maketitle
\thispagestyle{empty}
\pagestyle{empty}

\input{tex/0-abstract}

\section{Introduction}
\input{tex/1-intro}

\section{Related Work}
\input{tex/2-related_work}
\section{Method\label{sec:method}}
\input{tex/3-method}

\section{Experiments}
\input{tex/4-exp}
\section{Conclusion and Limitation}
\input{tex/5-conclusion}

{
\small
\bibliographystyle{IEEEtran}
\bibliography{references}
}

\end{document}

%% file: tex/0-abstract.tex
\begin{abstract}
Solving complex, long-horizon robotic manipulation tasks requires a deep understanding of physical interactions, reasoning about their long-term consequences, and precise high-level planning. Vision-Language Models (VLMs) offer a general perceive-reason-act framework for this goal. However, previous approaches using reflective planning to guide VLMs in correcting actions encounter significant limitations. These methods rely on inefficient and often inaccurate implicit learning of state-values from noisy foresight predictions, evaluate only a single greedy future, and suffer from substantial inference latency. To address these limitations, we propose a novel test-time computation framework that decouples state evaluation from action generation. This provides a more direct and fine-grained supervisory signal for robust decision-making. Our method explicitly models the advantage of an action plan, quantified by its reduction in distance to the goal, and uses a scalable critic to estimate. To address the stochastic nature of single-trajectory evaluation, we employ beam search to explore multiple future paths and aggregate them during decoding to model their expected long-term returns, leading to more robust action generation. Additionally, we introduce a lightweight, confidence-based trigger that allows for early exit when direct predictions are reliable, invoking reflection only when necessary. Extensive experiments on diverse, unseen multi-stage robotic manipulation tasks demonstrate a 24.6\% improvement in success rate over state-of-the-art baselines, while significantly reducing inference time by 56.5\%.
\end{abstract}

%% file: tex/1-intro.tex
Solving complex multi-stage manipulation tasks remains a core challenge in robotics~\cite{Du2025DynaGuideSD,Kroemer2019ARO,Cui2021TowardNL,bu2025univla}, especially when success hinges on comprehending sophisticated physical interactions and reasoning about their long-term implications~\cite{chen2025_value_learning}. These tasks often involve complex action sequences, where each step must consider physical constraints and potential consequences, thereby presenting substantial challenges to planning systems.
Some classical planning methods like Task and Motion Planning~\cite{Kaelbling2011HierarchicalTA}, offer a potentially viable solution but are constrained by their reliance on predefined symbolic representations and precise state estimation, limiting their use in partially observable environments.

Recent advancements in vision-language models (VLMs), leveraging internet-scale knowledge, have shown exceptional capabilities in processing visual scenes and natural language instructions for high-level task planning~\cite{Hurst2024GPT4oSC,ElKishky2024OpenAIOS,Comanici2025Gemini2P,Wang2024Qwen2VLEV,llava}. However, even state-of-the-art VLMs struggle with complex physical reasoning~\cite{Gao2023PhysicallyGV,Chow2025PhysBenchBA,Hu2023LookBY,yang2024adapt2reward}, especially concerning precise physical concepts and long-horizon planning. Previous approaches~\cite{Wang2022SelfConsistencyIC,Kwok2025RoboMonkeyST,Nakamoto2024SteeringYG} combining resampling with value functions for test-time action selection to improve planning capability, while they are hindered by suboptimal candidate action quality and inefficient iterative optimization, as all but the best-selected action are discarded.

Another line of work explores reflective planning. ReflectVLM~\cite{reflectvlm} is a two-stage test-time scaling framework comprising proposal and reflection stages. The proposal stage predicts an action conditioned on the current and goal visual states; the reflection stage then performs greedy foresight to predict a future visual state and appends the resulting image to the VLM to facilitate action revision. During post-training, the model is trained on concatenated observations of these visual states and supervised by expert actions. However, this framework implicitly learns state values without explicit value supervision, making it prone to mistaking task-irrelevant visual artifacts for meaningful progress. This entangles value learning with action generation, imposing excessive perceptual overhead and leading to an inefficient optimization path. Moreover, evaluating only a single greedy rollout rather than the expected long-term return introduces high-variance corrections that undermine decision-making robustness. Furthermore, the serial “reason–imagine–reason” workflow transforms single-pass inference into multiple sequential steps, substantially increasing latency.

To address these challenges, we introduce a novel test-time computation framework that decouples the evaluation of imagined future states from action generation.
We define the state-value as the distance to the goal state using an expert policy from the simulator and assess action plans by measuring the extent to which they reduce this distance, termed the advantage. A greater reduction signifies more promising actions, and we employ a critic to estimate these advantages during inference. This explicit value learning offers a more direct and fine-grained supervisory signal. 
In contrast to ReflectVLM, which implicitly learns value from noisy future visual observations across tasks, our post-training method employs value learning from a unified space as explicit learning signals, promoting inter-task knowledge sharing and enhancing generalization to novel environments.

To mitigate the stochasticity inherent in single-trajectory evaluation and promote knowledge sharing across trajectories, we introduce a multi-path reflection mechanism, as shown in Fig.~\ref{fig:framework}. This mechanism utilizes beam search to explore multiple multi-step future trajectories, aggregating them during decoding to model the expected long-term return, thereby enabling more robust next-step action generation. Unlike traditional sampling methods that select trajectories post-generation, our approach treats other trajectories within the set as complementary or contrasting inputs during decoding. By analyzing differences in output distribution, we can enhance or correct the current response within the generation process. This method functions as a test-time computation strategy, requiring no additional training.
To enhance inference efficiency, we then train a lightweight trigger that utilizes the model's hidden state to estimate its output confidence. We threshold these confidence scores to invoke the reflection stage only when necessary, thereby striking an optimal balance between efficiency and performance.

Experimental results demonstrate that our approach surpasses advanced VLMs, Monte Carlo Tree Search (MCTS), and the state-of-the-art solution, ReflectVLM~\cite{reflectvlm}. Notably, only with a single round of post-training, our approach surpasses ReflectVLM's success rate by 24.6\%, while using only 56.5\% of the inference time. This achievement underscores that value learning offers a more robust and effective learning signal, promoting rapid convergence and enhancing the decision-making capabilities of vision language models in complex manipulation tasks. Additionally, a straightforward output confidence assessment for early exit reduces overthinking without compromising performance.

Our contributions can be summarized as follows:
\begin{itemize}
    \item We introduce a value-guided reflective planning framework, demonstrating that explicit evaluation offers a more direct and nuanced learning signal, enabling the model to critically correct its actions.
    \item We propose a test-time computation framework that incorporates a multi-path reflection with a confidence-based early exit strategy, which enhances or refines outputs by aggregating them during decoding and achieves an equilibrium between success rate and efficiency.
    \item Experimental evaluations across 100 unseen tasks show that our method consistently surpasses state-of-the-art approaches in complex, multi-stage robotic manipulation tasks, while requiring less inference time.

\end{itemize}

%% file: tex/2-related_work.tex
\subsection{Reflection}

Recent research highlights the advantages of reflection mechanisms in Large Language Models (LLMs) and Vision Language Models (VLMs), which enhance outputs through iterative self-critique~\cite{Shinn2023ReflexionLA,renze2024self,Pan2024AutomaticallyCL,Madaan2023SelfRefineIR,Wang2022SelfConsistencyIC,Cheng2024VisionLanguageMC,Yu2024ExACTTA,Tang2025ReflectionWindowDT,wu2025leveraging,wu2022cikm}. For example, SELF-REFINE~\cite{Madaan2023SelfRefineIR} demonstrates that LLMs can improve initial outputs via self-refinement, while $R^{3}V$~\cite{Cheng2024VisionLanguageMC} and ExACT~\cite{Yu2024ExACTTA} propose reflecting on past failures that can improve search efficiency and generation quality. Reflection-Window Decoding~\cite{Tang2025ReflectionWindowDT} incorporates a sliding reflection window to refine outputs as the decoding proceeds.
However, these approaches primarily focus on past interactions rather than leveraging future foresight to optimize current outputs. ReflectVLM~\cite{reflectvlm} leverages a diffusion dynamics model to generate imagined future visual states for reflection, while this method reflects on a single future trajectory and its limited accuracy in evaluating future states leads to non-robust decisions. In contrast, we present a value-guided multi-path reflective planning framework that addresses these limitations and triggers reflection only when necessary to enhance efficiency.

\subsection{Test-Time Scaling for Robotic Manipulation Task}
A promising approach to enhancing the performance of pretrained robot policies is test-time scaling, which utilizes additional computational resources during deployment to refine actions. Current methodologies primarily fall into two paradigms. The first, ``generate-and-verify'', includes techniques such as V-GPS~\cite{Nakamoto2024SteeringYG}, RoboMonkey~\cite{Kwok2025RoboMonkeyST}, and FOREWARN~\cite{Wu2025FromFT}, which sample multiple candidate actions from a base policy and employ an external verifier (\textit{e.g.}, a value function or a VLM) to evaluate and select the optimal action independently. This paradigm suffers from suboptimal candidate outputs and information inefficiency as it discards sub-optimal candidates and their evaluation data. 
The second paradigm, ``model-based planning'', 
uses learned dynamics models to guide decision-making through forward simulation~\cite{Jain2025ASS,Du2025DynaGuideSD,Wu2025FromFT}. Although this allows for the generation of actions beyond initial proposals, these methods mainly rely on predicting a single, most probable future trajectory for each candidate action, making decision-making highly sensitive to model prediction errors and environmental variability. 
In contrast to these, we combine diffusion dynamic models with beam search to explore multiple future paths, enabling a comprehensive evaluation of candidate action plans and leveraging all input knowledge for output enhancement or correction during decoding.

%% file: tex/3-method.tex
\begin{figure*}[t]
    \centering 
    \includegraphics[width=0.86\textwidth]{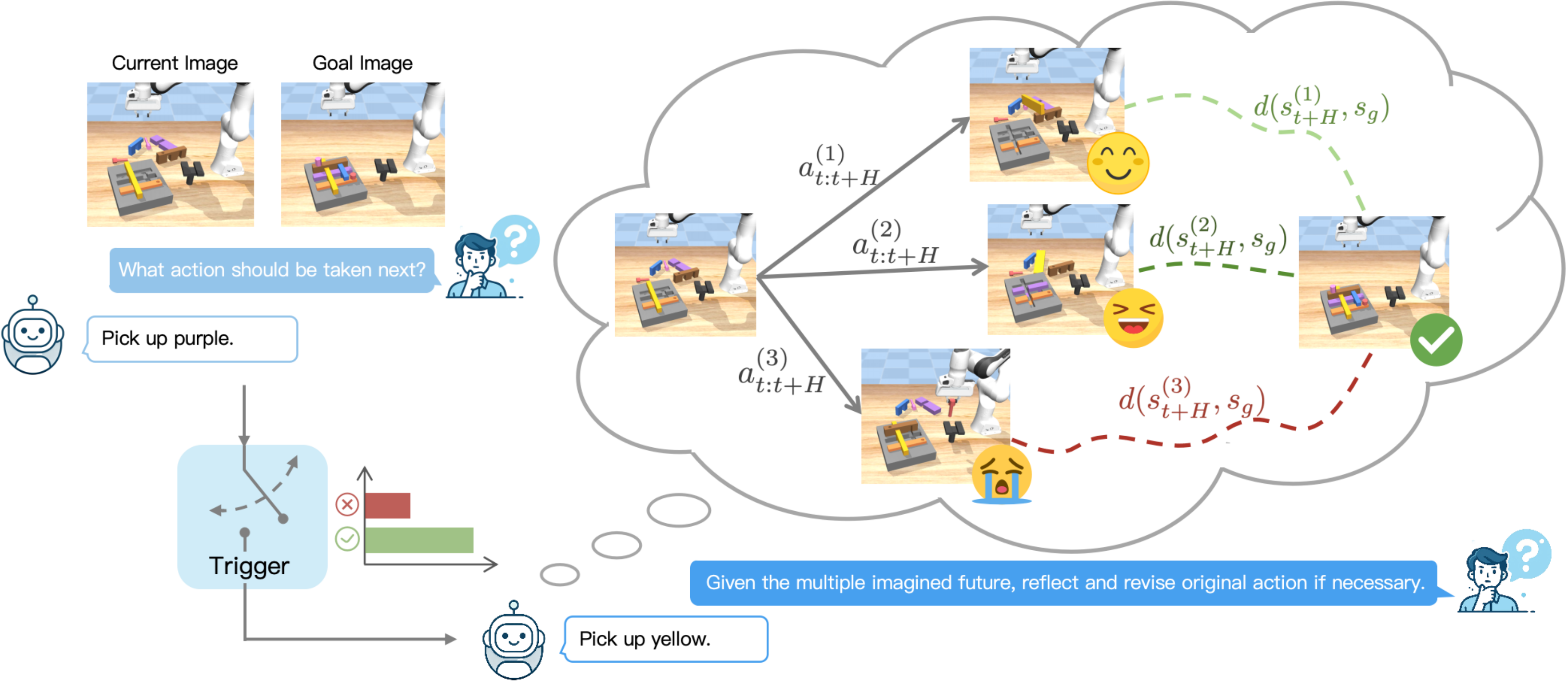}
    \vspace{-2pt}
    \caption{\textbf{Overview of our value-guided multi-path reflection framework}. Initially, the VLM policy proposes a candidate action list conditioned on the current and goal images ($\S$~\ref{sec:method-vlm}). The action with the highest probability (``pick up purple'') is treated as a preliminary plan and assessed by a confidence-based trigger ($\S$~\ref{sec:method-trigger}). If reflection is needed, a diffusion dynamics model imagines multiple future trajectories via beam search, guided by the candidate action list. A critic evaluates each of these imagined trajectories by estimating their distance reduction to the goal state, and this evaluation is incorporated as feedback into multiple distinct contexts. These inputs are then aggregated during the decoding stage, allowing the agent to reflect on the potential futures, revise its initial plan, and generate a more proper action (``pick up yellow'') ($\S$~\ref{sec:method-reflection}).}
    \label{fig:framework}
\end{figure*}

\subsection{Preliminaries and Problem Statement}
\label{sec:method-preliminary}
We formulate the multi-stage robotic manipulation planning problem as a Partially Observable Markov Decision Process (POMDP), characterized by the tuple $(\mathcal{S}, \mathcal{A}, \mathcal{T}, \mathcal{O}, \mathcal{Z})$. The state space $\mathcal{S}$ includes the complete physical state of the environment. The action space $\mathcal{A}$ consists of high-level manipulation actions: $\{\text{pick up, insert, reorient, put down}\} \times \{\text{objects}\}$, each with a failure rate $\epsilon$. Transition dynamics $\mathcal{T}(s_{t+1}|s_t, a_t)$ describe the physical interactions. RGB images form the observation space $\mathcal{O}$, and the observation model $\mathcal{Z}(o_t|s_t)$ maps states to images. We aim to develop a policy $\pi$ that generates action sequences to achieve a goal state $s_g$. The policy depends solely on image observations due to partial observability, represented as $\pi(a_t|I_t, I_g)$, where $I_t$ is the current observation and $I_g$ is the goal image. We implement this policy as a Vision-Language Model (VLM) agent, $\pi_{\text{VLM}}$, which takes multi-modal inputs of images and text and outputs action primitives in text form.

Our framework consists of a pre-training phase and a post-training phase. The post-training phase uses interactive imitation learning~\cite{Ross2010ARO,Kelly2018HGDAggerII}, where a policy is learned by interacting with the environment and receiving real-time expert guidance. We assume access to an expert policy $\pi_E$ that provides near-optimal actions $a^* = \pi_E(s)$ for any state $s$ during training. In this study, we implement the expert policy with full access to the environment's state to generate optimal actions, while the VLM policy only uses image observations.

\subsection{Value-Guided VLM Policy Post-Training}
\label{sec:method-vlm}
VLMs trained via imitation learning inevitably inherit the covariant shift defects.  To address this, we employ an interactive post-training phase like DAgger~\cite{Ross2010ARO,Kelly2018HGDAggerII}, following ReflectVLM~\cite{reflectvlm}. This allows the VLM to learn prospective reflection and self-correct its policy via environmental interaction.
We iteratively enhance the VLM policy by rolling it out in the environment to collect data for finetuning. As detailed in Alg.~\ref{alg:training}, at each timestep, the VLM is prompted with the goal image and the current image to generate a learner action $a_t^L$. We also generate an expert action $a_t^*$ from the expert policy.  These pairs, $((I_g, I_t), a_t^*)$, are incorporated into the dataset for further finetuning. To ensure effective convergence, we execute the learner action $a_t^L$ with probability $p$ and the expert action $a_t^*$ with probability $1-p$. 

Similar to ReflectVLM~\cite{reflectvlm}, we can incorporate the future image $I_{t+H}$ (obtained after executing the action sequence $a_{t:t+H-1}$) into the reflective context at time step $t$ to generate training data for reflection. However, this paradigm, which allows the VLM to implicitly evaluate, can result in inefficient learning due to vague and subjective supervision signals. Moreover, these images often contain task-irrelevant visual noise, leading the model to learn incorrect or false associations. These issues may increase the risk of overfitting on limited post-training data, hindering generalization to new tasks. To address these challenges, we propose explicitly quantifying the cumulative return of planned actions.

Previous research~\cite{Hejna2023DistanceWS,Ma2022VIPTU} has proved a precise one-to-one correspondence between the cumulative return and its distance to the goal. In this context, the cumulative discounted return for a policy reaching the goal in \( m \) steps is a deterministic function solely dependent on \( m \) and the discount factor. This finding implies that estimating the cumulative return of planned actions is equivalent to predicting their distance to the goal state using the oracle action at each step. Thus, we treat value as the distance from the current state to the goal.
We represent the distance from state \( s \) to goal \( s_g \) as \( d(s, s_g) \). When evaluating two action plans, \( a^{(1)} \) and \( a^{(2)} \), which result in states \( s^{(1)} \) and \( s^{(2)} \) respectively, if \( d(s^{(1)}, s_g) < d(s^{(2)}, s_g) \), indicating that \( s^{(1)} \) is closer to the goal, action plan \( a^{(1)} \) is preferred. 
To achieve a unified metric space, we evaluate the action plans according to their reduction in distance to the goal, which we refer to as the advantage 
$\Delta d_t^{(H)}\triangleq d(s_t,s_g)-d(s_{t+H},s_g)$, that we can easily get from the simulator with an expert policy. $\Delta d_t^{(H)} < 0$ means proposed action plans lead to a terrible future state, which is far from the goal state. We template this distance reduction as language feedback for reflection.
To generate training data for reflection, we can simply relabel a trajectory after it is terminated by appending $\Delta d_t^{(H)}$. We then use a critic trained to estimate it during inference.
We generate two types of question-answering examples through interaction with the environment. The first involves predicting an optimal action based on images of the goal and current state. The second requires reflecting on and revising an initial action sequence by considering additional distance reductions. We denote $\mathbf{x}^{p}$ and $\mathbf{x}^{r}$ as the two types of prompts for these question-answering examples. 
The VLM is trained to generate actions aligned with expert actions in the dataset with a cross-entropy loss:
\begin{equation}
\scalebox{0.89}{$
\begin{aligned}
 & \min_{\pi_{\mathrm{VLM}}}\mathbb{E}_{\mathcal{D}}\Big[\mathcal{L}_{\mathrm{CE}}(\pi_{\mathrm{VLM}}^{\mathrm{propose}}(a_{t}|I_{g},I_{t},\mathbf{x}^{p}),a_{t}^{*}) \\
 & +\mathcal{L}_{\mathrm{CE}}(\pi_{\mathrm{VLM}}^{\mathrm{reflect}}(a_{t}|I_{g},I_{t},\Delta d_t^{(H)},a_{t:t+H-1},\mathbf{x}^{r}),a_{t}^{*})\Big].
\end{aligned}
$}
\end{equation}

\begin{algorithm}[t]
\caption{Value-Guided Interactive VLM Post-Training}
\label{alg:training}
\begin{algorithmic}[1]
\REQUIRE {initial state distribution $\rho_0$, goal state distribution $\rho_g$, number of iterations $K$, number of trajectories per iteration $N$, episode length $T$, imagination horizon $H$, expert policy $\pi_E$, expert demonstrations $\mathcal{D}^*$}
\STATE train base policy $\pi_\text{VLM}$ on $\mathcal{D}^*$
\STATE $\mathcal{D}\gets\mathcal{D}^*$
\FOR {$i \gets 1$ to $K$}
    \STATE $\mathcal{D}_i\gets\emptyset$
    \STATE {rollout out policy $\pi_\text{VLM}$ to collect data $\mathcal{D}_{i}$}
    \FOR {$n \gets 1$ to $N$}
        \STATE $s_0\sim\rho_0$; $I_0\gets\mathcal{Z}(s_0)$
        \STATE $s_g\sim\rho_g$; $I_g\gets\mathcal{Z}(s_g)$
        \FOR {$t \gets 0$ to $T-1$}
            \STATE $a_t^{\dag}\sim\pi_\text{VLM}(I_g, I_t)$; $a_t^* \sim \pi_E(s_g, s_t)$
            \STATE $a_t\gets a_t^{\dag}$ \textbf{if} ${\tt random()} <p$ \textbf{else} $a_t^*$
            \STATE $s_{t+1}\gets\mathcal{T}(s_t, a_t)$; $I_{t+1}\gets\mathcal{Z}(s_{t+1})$
        \ENDFOR
        \STATE $\mathcal{D}_i\!\gets\!\mathcal{D}_i\cup\{((I_g, I_t), a_t^*)\}_{0\le t<T}$
        \STATE {{$\mathcal{D}_i\!\gets\!\mathcal{D}_i\cup\{((I_g, I_t, \Delta d_t^{(H)}, a_{t:t+H-1}), a_t^*)\}_{0\le t<T}$}}
    \ENDFOR
    \STATE {$\mathcal{D}\gets \mathcal{D}\cup\mathcal{D}_{i}$}
    \STATE {finetune $\pi_\text{VLM}$ on $\mathcal{D}$}
\ENDFOR
\end{algorithmic}

\end{algorithm}

\subsection{Reflection from Multiple Future Trajectories}
\label{sec:method-reflection}
\begin{figure}[t]
    \centering 
    \includegraphics[width=0.5\textwidth]{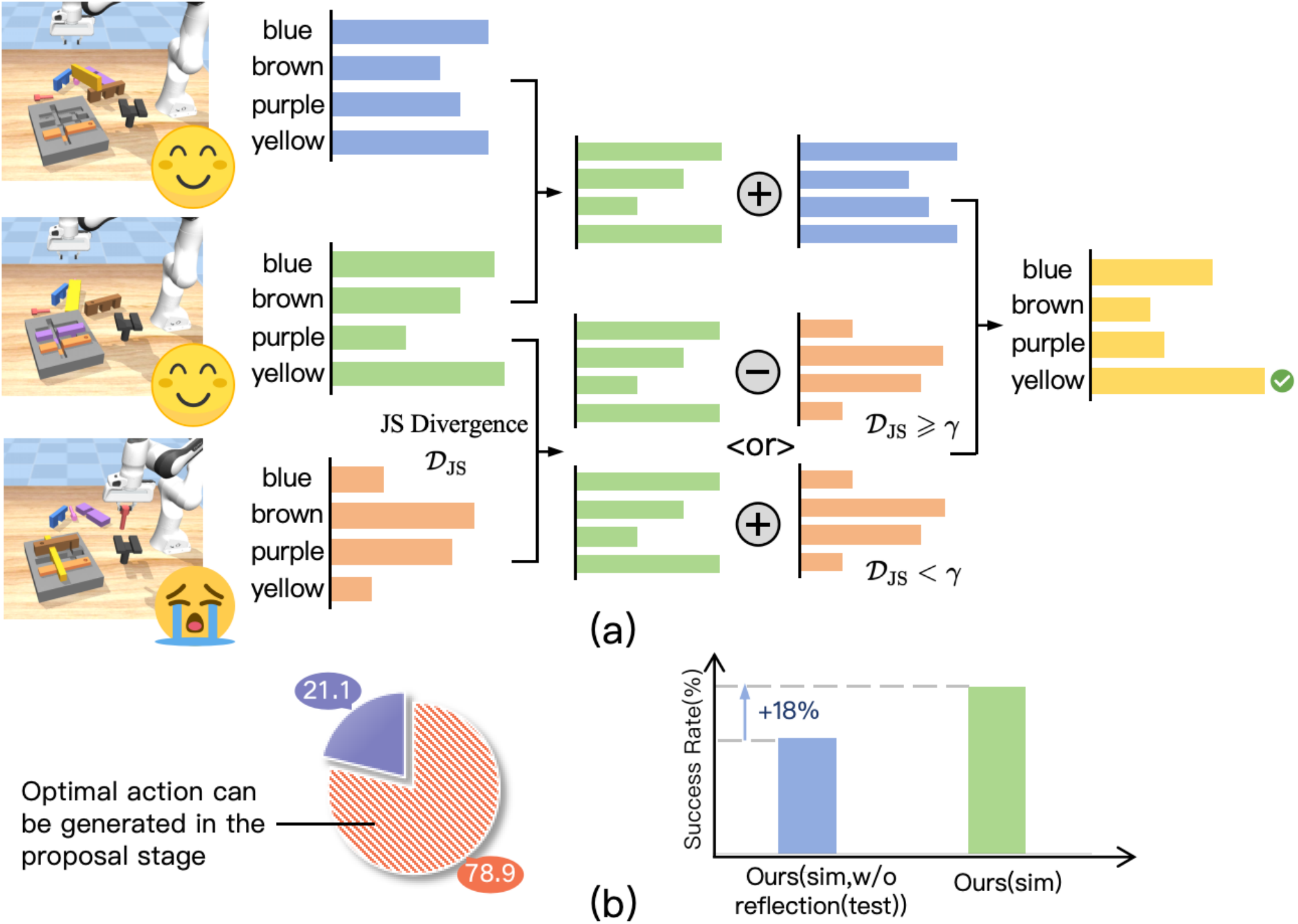}
    \caption{(a) \textbf{Overview of the multi-path reflection mechanism during decoding}. Output distributions from multiple inputs based on distinct imagined futures are dynamically combined using complementary or contrastive decoding based on their Jensen-Shannon Divergence to enhance or correct output. (b) A substantial 78.9\% of optimal actions are produced in the initial proposal stage, while the reflection stage is also essential for more challenging decisions, leading to an 18\% increase in overall success.} 
    \label{fig:decoding&early-exit}
\end{figure}

Reflective planning evaluates and optimizes actions by analyzing individual future paths. However, relying on a single greedy path can lead to uncertain and incorrect answers. To address this, in Fig.~\ref{fig:decoding&early-exit}, we propose a multi-path reflective planning framework for the inference phase. It is a test-time scaling framework that performs parallel forward searches and aggregation of these multiple future trajectories of length $H$, using a learned dynamic model for internal trial and error to find the best action plan. We use a fixed-length beam search to generate these trajectories based on candidate actions generated from the proposal phase. At each search step, we retain a select number of optimal candidates and continue the search from these. We leverage the critic to estimate the advantage of these multiple future paths and add into the prompt as language feedback. Additionally, we parallelize the evaluation and generation of scenarios at each step, avoiding the extra time costs of a global search.
Specifically, we prompt the VLM to generate $K$ independent output probability distributions  $p_{i}^{k}$, conditioned on \( I_t \), \( I_g \), \(\Delta d_{t,k}^{(H)}\) and \(a^k_{t:t+H-1}\), which is expressed as follows:
 \begin{equation}
 \begin{aligned}
\scalebox{0.89}{
  $\displaystyle{
p_{i}^{k} = \mathrm{Softmax}\left[\pi_\theta(\eta_{i}|I_{t},I_{g},\Delta d_{t,k}^{(H)},a_{t:t+H-1}^{k},\mathbf{x}^{r},\eta_{<i})\right],
    }$
    }
\end{aligned}
\end{equation}
where $K$ denotes the number of imagined future trajectory, and \(\Delta d_{t,k}^{(H)}\) represents the distance reduction estimation resulting from $k$-th future trajectory and $\pi$ denotes VLM policy parameterized by $\theta$.
In order to aggregate them and output a final action, a simple solution is to leverage some sample method after action generation, like majority voting~\cite{Wang2022SelfConsistencyIC} and Best-of-N~\cite{Lightman2023LetsVS,Zhang2024GenerativeVR}. However, these approaches are often constrained by the quality of candidate outputs and suffer from inefficient information utilization. They typically select only the best option and discard others, thereby failing to facilitate knowledge transfer across different output distributions from various future path inputs. Inspired by some previous work~\cite{woo2024ritual,Leng2023MitigatingOH,Zhang2025SelfCorrectingDW,Kim2024CODECS}, we propose an aggregation strategy during the decoding phase as shown in Fig.~\ref{fig:decoding&early-exit}(a). We treat other future paths within the set as complementary or contrasting inputs during decoding. By analyzing differences in their output distribution, we can constantly enhance or correct the current response in generation process.

Specifically, based on their distance reduction \(\Delta d_{t,k}^{(H)}\), these $K$ parallel streams are stratified into three distinct sets: a baseline set $\mathcal{S}_{base}$ composed of the top-$N$ ranked trajectories, a promising reference set $\mathcal{S}_{ref}^{p}$, and a suboptimal reference set $\mathcal{S}_{ref}^{n}$, with the latter two partitioned from the remaining trajectories by an advantage threshold $\sigma$. The core of our decoding mechanism lies in dynamically enhancing or correcting the baseline prediction, using predictions from the reference sets. For token predictions from the promising reference set, we employ complementary decoding~\cite{woo2024ritual} to enhance their consensus:
$p_i^{k,j} = \mathrm{Softmax}\big[f^{k} + \alpha_{1}f^{j}\big]$,
where $k \in \mathcal{S}_{base}$, $j \in \mathcal{S}_{ref}^{p}$, and $f^k=\pi_\theta(\eta_{i}|I_{t},I_{g},\Delta d_{t,k}^{(H)},a_{t:t+H-1}^{k},\mathbf{x}^{r},\eta_{<i})$.
For those from the suboptimal reference set, we utilize a hybrid strategy based on the Jensen-Shannon Divergence $\mathcal{D}_{\mathrm{JS}}$ between their predictions and the baseline prediction; complementary decoding~\cite{woo2024ritual} is applied in cases of low divergence, which means their prediction distributions are similar, while contrastive decoding~\cite{Leng2023MitigatingOH,Kim2024CODECS} is used in cases of high divergence to suppress potential errors.
To implement this, we introduce a distance threshold $\gamma$ and develop two corresponding decoding approaches as follows:
\begin{equation}
\scalebox{0.89}{
  $\displaystyle{
    p_i^{k,m} = \begin{cases}\mathrm{Softmax}\left[ f^{k} + \alpha_{1}f^{j}\right],&\mathrm{if}\; \mathcal{D}_{\mathrm{JS}}<\gamma; \\
    \mathrm{Softmax}\left[ (1+\alpha_2)f^{k} - \alpha_{2}f^{j}\right],&\mathrm{if}\;  \mathcal{D}_{\mathrm{JS}}\ge \gamma;\end{cases}
    }$
    }
\end{equation}
Finally, the next token prediction $\eta_i$ is computed by averaging their probabilities:
\begin{equation}
\scalebox{0.89}{
  $\displaystyle{
\eta_i ~\sim \frac{1}{|\mathcal{S}_{base}|} \sum_{k \in \mathcal{S}_{base}} \left( \frac{1}{|\mathcal{S}_{ref}^{p} \cup \mathcal{S}_{ref}^{n}|} \sum_{l \in \mathcal{S}_{ref}^{p} \cup \mathcal{S}_{ref}^{n}} p_i^{k,l} \right)
    }.$
    }
\end{equation}

\subsection{Confidence-based Early-Exit for Efficient Planning}
\label{sec:method-trigger}
Our reflective planning offers the advantage of exploring multiple paths to incrementally approach the correct solution by assessing the cumulative return of various action plans. However, this method can be inefficient, as reflective reasoning may continue even after generating the optimal action during the proposal phase. 
As shown in Fig.~\ref{fig:decoding&early-exit}(b), we illustrate that 78.9\% of optimal actions are generated during the proposal stage, enabling direct responses without necessitating reflective planning. Nevertheless, the minority of actions requiring adjustment during the reflection stage are pivotal for achieving task success, resulting in an 18\% increase in success rate compared to relying exclusively on the proposal stage for reasoning. These findings raise a critical question: how can we develop a strategy to adaptively determine whether to initiate the reflection stage?

Thus, we propose leveraging the model's hidden states to decide whether to trigger the reflection stage. Ideally, when the model effectively utilizes encoded correctness information, it should employ an optimal strategy by halting inference when the confidence in the proposed action is sufficiently high. This approach is essential to prevent overthinking and ensure that the reflection phase is triggered only when needed. To implement this, we train a two-layer MLP binary classifier as a trigger, employing a confidence-based early exit strategy by establishing a confidence score threshold. 
We roll out the policy in the environment to collect training data $\mathcal{D}_{tr}$, and at each step of the trajectory, we extract the last layer hidden state at the last token position of the proposal phase output, denoted as $e_i$. If the proposed action matches the expert action, the label is set to 1; otherwise, it is set to 0. We denote $N_{tr}$ as the length of dataset $\mathcal{D}_{tr}$. To address the class imbalance in the dataset collected from the training task, where the majority of samples represent correct actions, we utilize a weighted binary cross-entropy loss:
\begin{equation}
\scalebox{0.89}{
  $\displaystyle{
p_{i}^{tr}  =\sigma(\operatorname{ReLU}(e_i\mathbf{W}_1+\mathbf{b}_1)\mathbf{W}_2+b_2)
    },$
    }
\end{equation}
\begin{equation}
\scalebox{0.88}{
  $\displaystyle{
\mathcal{L}_{BCE}  =-\frac{1}{N_{tr}}\sum_{i=1}^{N_{tr}}\left(w\alpha_{3} y_i\log p^{tr}_i+(1-y_i)\log(1-p^{tr}_i)\right)
    }.$
    }
\end{equation}

\subsection{Planning Framework}
\label{sec:method-planning}
As illustrated in Fig.~\ref{fig:framework}, our planning framework comprises four components: a VLM policy, a diffusion dynamics model, a critic, and a trigger. The diffusion dynamics model predicts future visual observations conditioned on the current observation and a candidate action sequence. The critic estimates the advantage of each candidate plan given images of the current, predicted, and goal states. At each time step, planning proceeds in two phases—proposal and reflection—both executed by the same VLM policy with distinct prompt templates; a trigger gates entry into the reflection phase.

During the proposal phase, the VLM takes the current state $I_t$ and goal state $I_g$ and outputs a candidate action sequence. The last-layer hidden state of the last output token with the highest probability score is passed to a trigger that estimates an early-exit probability. If this probability is below a preset threshold, the reflection phase is invoked to self-correct the plan via forward search; otherwise, the procedure exits early.
During the reflection phase, we run \(H\) iterations of action proposal and diffusion-based image prediction, to forecast an \(H\)-step future (planning horizon \(H\)). 
We use a fixed-width beam search of width \(W\), expanding at each step only the top-\(W\) actions by probability scores, which yields multiple \(H\)-step candidate trajectories. We leverage critic to estimate each trajectory’s advantage (goal–distance reduction). We verbalize this signal into a concise language description as long-term return feedback and append it together with the current and goal states to the VLM input. To aggregate outputs across multiple future paths, we apply complementary or contrastive decoding, to enhance or correct the output.

%% file: tex/4-exp.tex
\subsection{Experiment Settings}
\begin{figure}[t]
    \centering 
    \includegraphics[width=0.45\textwidth]{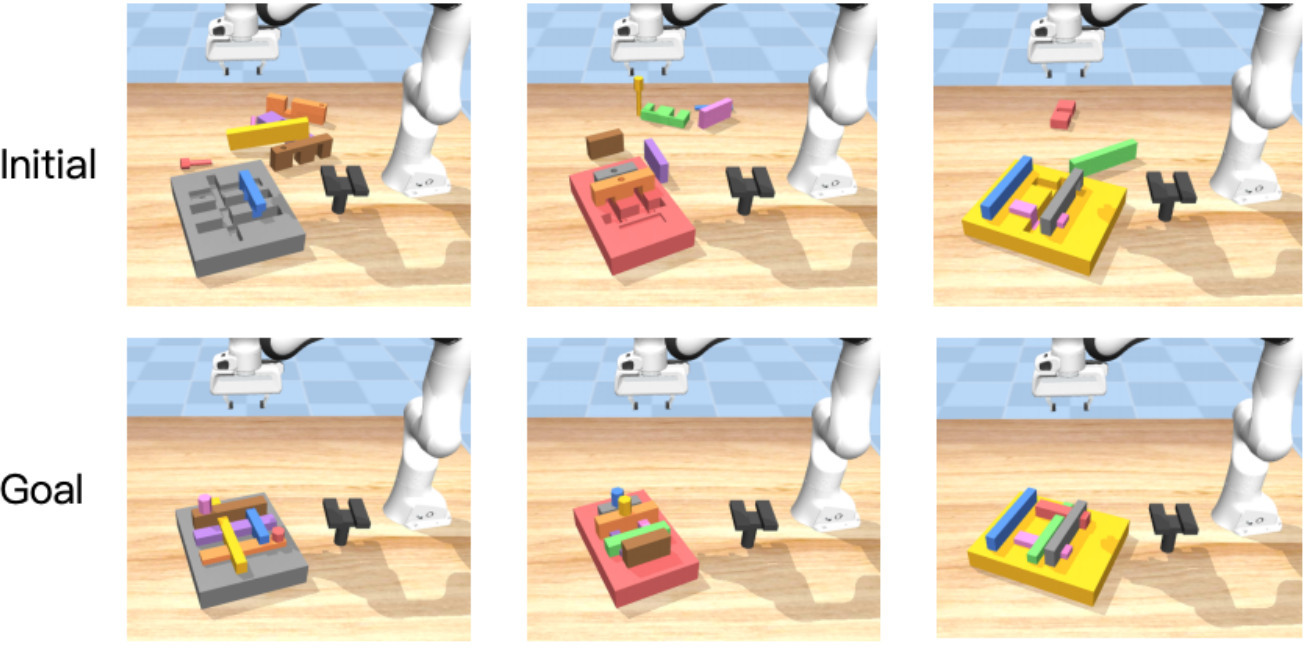}
    \caption{\textbf{Task examples.}}
    \label{fig:task examples}
\end{figure}
\begin{figure*}[t]
    \centering 
    \includegraphics[width=0.88\textwidth]{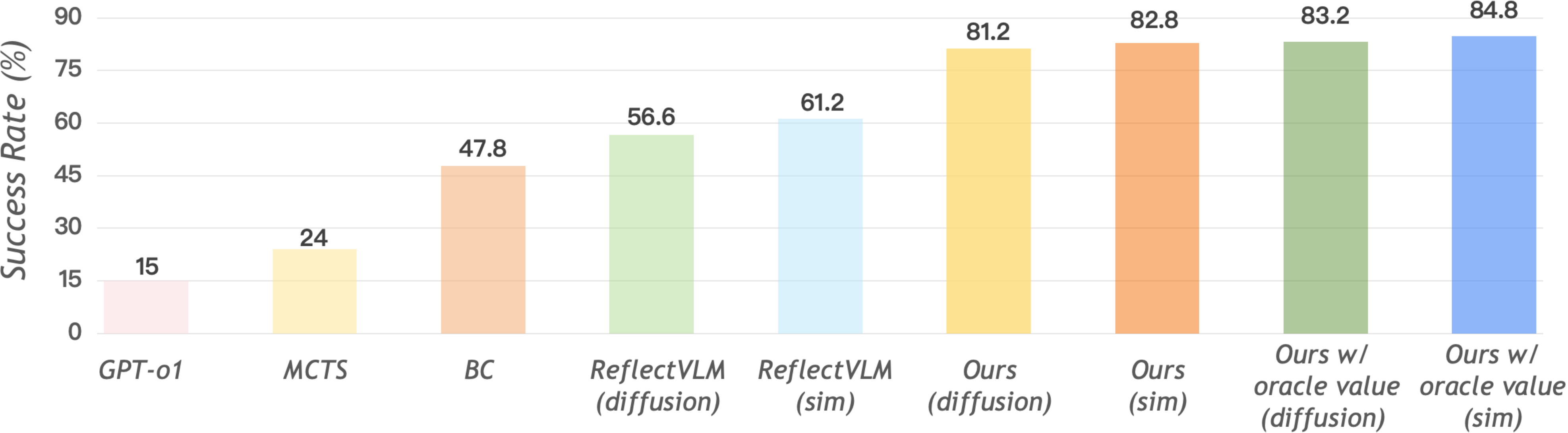}
    \caption{\textbf{Comparison of success rates on 100 unseen long-horizon manipulation tasks.} Our value-guided reflective planning framework, in both its simulation-based (sim) and diffusion-based variants, significantly outperforms all baselines, including traditional methods like MCTS and Behavioral Cloning (BC) and the prior state-of-the-art, ReflectVLM. For a fair comparison, both our method and ReflectVLM are evaluated under only a single round of post-training. The ``Ours w/ oracle value'' results demonstrate the upper-bound performance of our framework with perfect value estimation.} 
    \label{fig:comparative_exp}
\end{figure*}
\textbf{Environment Settings.}
Following ReflectVLM~\cite{reflectvlm}, we introduce a suite of multi-stage, long-horizon manipulation tasks that require understanding physical interactions and reasoning about long-term action sequences. The tasks begin with a board and randomly placed pieces on a table, aiming to fully assemble the board by sequentially inserting the pieces. Initial and goal states are illustrated in Fig.~\ref{fig:task examples}. Many tasks involve interlocking pieces that must be inserted in a specific order, necessitating strategic object selection and interaction inference. This interlocking feature requires the agent to replan, recovering from errors due to prior mistakes or poor initialization. We focus on high-level planning, defining actions as ``\texttt{[act] [obj]}'' where $\texttt{[act]} \in \{\texttt{pick up, insert, reorient, put down}\}$ and \texttt{[obj]} is the object to manipulate. ``\texttt{reorient}'' adjusts an object for insertion. Each action is executed by a rule-based script controller, but using other low-level controllers, such as learning-based policies like behavior cloning, is possible. 

\textbf{Task Settings.} To evaluate the generalization capability, we create two distinct task sets: a training set and an evaluation set with unseen configurations. The evaluation tasks are designed to test generalization across diverse object configurations, colors, and spatial arrangements. The training set includes 1,000 tasks, each with five randomized initial arrangements, used for pre-training the VLM policy. During post-training, 1000 tasks are sampled to further train the VLM policy using the reflection mechanism. The evaluation set consists of 100 tasks not included in the training set. 
\textbf{Training Settings}. During the policy pre-training phase, we utilize the expert policy to provide action labels, then finetune an LLaVa-1.5-13B model~\cite{liu2023visual,llava} with standard supervised learning loss. This pre-training uses 5,000 expert demonstrations (1,000 unique tasks × 5 initial configurations per task). In the post-training phase, we use the same expert policy to further train the VLM policy as shown in Alg.~\ref{alg:training}. Unlike ReflectVLM, we only conduct one round of post-training, and we collect 1k trajectories by rolling out the VLM policy in the environment to generate examples for fine-tuning. During collecting post-training data, beyond the collection of \(\{I_t, I_g, I_{t+H}, a_{t:t+H}\}\), we also acquired distance reduction $\Delta d_t^{(H)}$, hidden states $e_t$, and their labels  $y_t$. We construct datasets \(D_{\text{critic}}\) and \(D_{\text{trigger}}\) using re-annotated post-training data and expert data. For training, the datasets are randomly partitioned into training and validation sets (9:1). Following ReflectVLM, we set $H$ as 5. The critic is built upon a ResNet34 backbone and trained with MSE loss, while the trigger is a two-layer MLP with a 64-dimensional hidden layer. These two models are trained for at most 100 epochs with a batch size of 64 using the AdamW optimizer, with validation accuracy as the early stopping criterion. Both of them are scalable and can be further developed to enhance their performance in the future. We use the pre-trained InstructPix2Pix model~\cite{Brooks2022InstructPix2PixLT} as our diffusion dynamic model, further training it in two phases.

\subsection{Results}
\begin{figure*}[t]
    \centering 
    \includegraphics[width=0.98\textwidth]{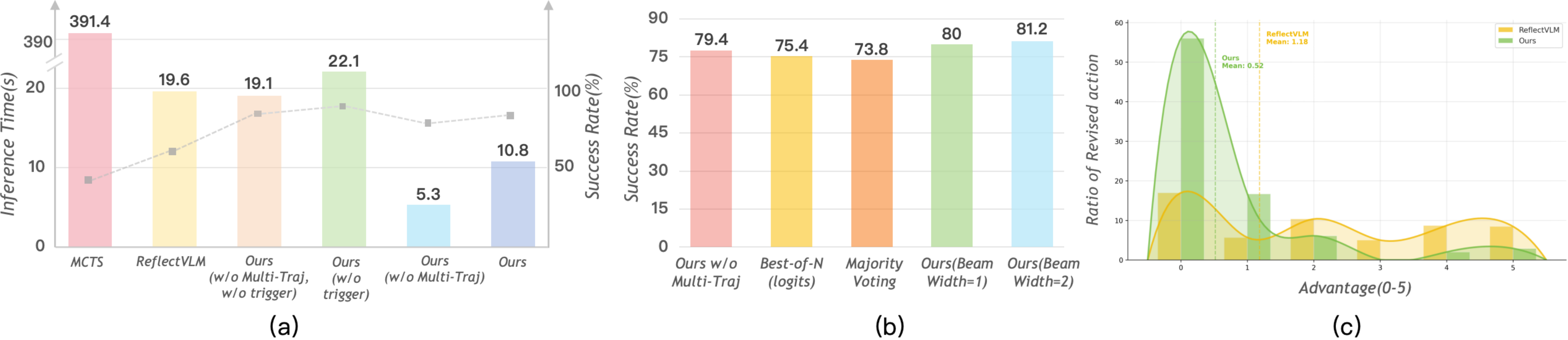}
    \caption{(a) \textbf{Inference time cost comparison}. Our method significantly outperforms ReflectVLM~\cite{reflectvlm}, reducing inference time by over 45\% via the confidence-based early-exit strategy. (b) \textbf{Ablation study on multi-path aggregation strategies}. Our aggregation mechanism during decoding outperforms both the single-path baseline and traditional post-hoc selection methods, such as Best-of-N and Majority Voting. (c) \textbf{Qualitative analysis of the advantage distribution for revised actions.} Our method's reflection is highly precise, primarily targeting low-advantage actions (peaking near 0). In contrast, the baseline's reflection is indiscriminate and less likely to trigger necessary revisions.} 
    \label{fig:exp2}
\end{figure*}
\textbf{Comparative Experiments}.
To comprehensively evaluate the performance of our method, we conduct experiments on complex robotic manipulation tasks requiring multi-stage, long-horizon operation. Our evaluation is designed to measure the model's generalization capabilities in 100 unseen tasks across five configurations. In Fig.~\ref{fig:comparative_exp}, we show the success rate of representative baseline methods with our method.
We select several methods as baselines for a comprehensive comparison. First, we evaluate the ability of an advanced Vision-Language Model, 
GPT-o1~\cite{ElKishky2024OpenAIOS}, to solve these complex manipulation tasks in a zero-shot setting without any task-specific training. To compare with model-based planning approaches, we implement a VLM-based Monte Carlo Tree Search (MCTS), which uses a pretrained VLM as its base policy for generating candidate actions when expanding tree nodes, with value estimation provided by the expert policy from the simulator. We also train a fundamental imitation learning model, 
Behavioral Cloning (BC), via standard supervised learning on 5,000 expert demonstrations without advanced planning or reflection mechanisms. Finally, we conduct a comparison with ReflectVLM, an advanced method that introduces a post-training strategy with a reflection mechanism to optimize VLM policies. This method also includes two variants for its inference phase: one utilizing the simulator (sim) for lookahead planning and another using a diffusion model (diffusion) to predict future states.

Our method demonstrates a significant improvement in success rates over all baseline approaches. Traditional planning methods, such as Zero-Shot VLM (15\%), MCTS (24\%), and Behavioral Cloning (47.8\%), under-perform, highlighting the necessity for advanced planning and physical reasoning capabilities beyond simple imitation or zero-shot inference. 
For a more direct comparison with the current SOTA, we evaluate both our method and ReflectVLM using a single round of post-training to ensure fairness. In this setting, ReflectVLM achieves success rates of 56.6\% using diffusion and 61.2\% with a simulator. ReflectVLM's performance is limited by its dependence on implicitly learning state-values from noisy visual predictions and the stochastic nature of evaluating a single, greedy future trajectory. In contrast, our method achieves significantly higher success rates of 81.2\% with a diffusion model and 82.8\% using a simulator using a single round post-training. 
Crucially, these single-round results are already comparable to those achieved by ReflectVLM after three full iterations of data collection and post-training, highlighting the superior data efficiency of our framework.
This substantial advantage is attributed to two key innovations: an explicit value critic that directly models an action's ``advantage'' by its goal-distance reduction, and a multi-trajectory reflection that aggregates futures during decoding to mitigate the stochasticity of single-path evaluation.
Furthermore, the ``Ours w/ oracle value'' variants, which utilize a ground-truth oracle for value estimation instead of our learned critic, achieve the highest success rates of 83.2\% (diffusion) and 84.8\% (sim). These results underscore the upper-bound potential of our framework, demonstrating its capabilities when supplied with perfect value information.

\textbf{Inference Time Cost.}
In addition to superior success rates, we also evaluate the computational efficiency of our method. As shown in Fig.~\ref{fig:exp2}(a), the MCTS baseline was impractically slow, requiring 391.4 seconds per step, highlighting the inefficiency of traditional planning methods. More importantly, we demonstrate a significant speedup over the ReflectVLM baseline. ReflectVLM’s serial ``reason-imagine-reason'' workflow results in a notable latency of 19.6s per step. In contrast, our standard method is 45\% faster, requiring only 10.8s. 
This substantial efficiency gain is primarily due to our confidence-based early exit strategy. Our ablation study further validates the efficacy of this trigger mechanism: disabling it and enforcing reflection at every step increases inference time to 19.1s and 22.1s. These results demonstrate that our triggered approach effectively reduces inference time by over 45\%, while maintaining minimal impact on the final success rate, achieving an optimal balance between performance and efficiency.

\textbf{Ablation Study.}
To validate the effectiveness of our proposed multi-path reflection mechanism, we conduct an ablation study comparing it against several alternative aggregation strategies.
As shown in Fig.~\ref{fig:exp2}(b), our model using only a single reflection trajectory ``Ours w/o Multi-Traj'' achieves a success rate of 79.4\%, which serves as a strong baseline. Interestingly, naive aggregation methods that operate on fully generated candidate actions perform significantly worse. Both Best-of-N using probability score and Majority Voting degrade performance, with success rates of 75.4\% and 73.8\%, respectively. This result validates our hypothesis that such post-hoc selection methods are suboptimal, as they fail to achieve knowledge transfer between different potential futures and discard valuable information from all but the chosen trajectory.
In stark contrast, our proposed method, which treats multiple trajectories as complementary or contrasting inputs during the decoding process, yields significant performance gains. Even with a beam width of one, our decoding strategy improves the success rate to 80\%. 
By increasing the width of the beam search, 
the success rate further increases to 81.2\%. This demonstrates the dual benefit of our approach: the decoding strategy provides a more robust way to generate actions and this benefit is amplified when it can draw upon a richer set of imagined futures. 

\textbf{Qualitative Analysis.}
To evaluate the precision of our value estimation framework, we analyze the distribution of the advantage for actions that the model chooses to revise, as shown in Fig.~\ref{fig:exp2}(c). Advantage is defined as the reduction in distance to the goal after an action, with a higher value indicating a better action. Our method exhibits a highly concentrated distribution with a sharp peak centered around an advantage of 0, and the mean advantage of revised actions is only 0.52. This demonstrates that our reflection is highly precise, overwhelmingly targeting actions that offer little progress. This precision is a direct result of our explicit value critic, which accurately assesses an action plan's value. In contrast, the distribution for the ReflectVLM baseline is flat and spread out. This indicates that ReflectVLM's reflection is far less discriminating, frequently revising actions of all quality levels, including those that are already highly advantageous, leading to inefficient ``overthinking.'' This behavior aligns with the known limitations of its implicit visual evaluation, which struggles to reliably judge a plan's quality. Furthermore, the total ratio of revised actions is visibly lower for ReflectVLM, indicating that it may not revise actions, even when required. This analysis confirms the superior precision of our framework; it focuses reflection effort on correcting genuinely suboptimal actions, while the baseline wastes computation on indiscriminate revisions.

%% file: tex/5-conclusion.tex
We propose a value-guided, multi-path reflective planning framework that decouples evaluation of imagined futures from action generation. Advantage is defined as the reduction in goal distance, and a scalable critic is trained to estimate this signal for explicit value learning. During decoding, multi-path reflection aggregates several imagined futures to approximate long-term returns and iteratively refine the plan. A lightweight trigger enables early exit when confidence is sufficient, substantially reducing test-time computation while maintaining success rates. Extensive experiments on multi-stage, long-horizon manipulation tasks show that our approach outperforms state-of-the-art baselines and yields more reliable next-step decisions with lower inference cost.

Despite strong performance and efficiency, real-robot deployment remains difficult due to the expense of collecting high-quality interaction data and ground-truth labels, as well as sim-to-real discrepancies in dynamics and contact-rich interactions. Future work can develop a hierarchical system that integrates high-level VLM planning with low-level Vision-Language-Action (VLA) control and supports closed-loop self-improvement for complex, multi-stage manipulation in real robot settings.